# VQA support to Arabic Language Learning Educational Tool

Khaled Bachir Delassi[1], Lakhdar Zeggane[1], Hadda Cherroun[1],
Abdelhamid Haouhat[1], Kaoutar Bouzouad[2]
[1]LIM Lab, Amar Telidji University, Laghouat, Algeria
{k.delassi.mi, l.zeggane.mi, hadda_cherroun, a.haouhat}@lagh-univ.dz
[2]Computer Science Dept., USTHB, Algiers, Algeria
ktr.bouzouad@usthb.dz

*Abstract*—We address the problem of scarcity of educational Arabic Language Learning tools that advocates modern pedagogical models such active learning which ensures language proficiency. In fact, we investigate the design and evaluation of an AI-powered educational tool designed to enhance Arabic language learning for non-native speakers with beginner-to-intermediate proficiency level. The tool leverages advanced AI models to generate interactive visual quizzes, deploying Visual Question Answering as the primary activity. Adopting a constructivist learning approach, the system encourages active learning through real-life visual quizzes, and image-based questions that focus on improving vocabulary, grammar, and comprehension. The system integrates Vision-Language Pretraining models to generate contextually relevant image description from which Large Language Model generate assignments based on customized Arabic language Learning quizzes thanks to prompting.

The effectiveness of the tool is evaluated through a manual annotated benchmark consisting of 1266 real-life visual quizzes, with human participants providing feedback. The results show a suitable accuracy rates, validating the tool's potential to bridge the gap in Arabic language education and highlighting the tool's promise as a reliable, AI-powered resource for Arabic learners, offering personalized and interactive learning experiences.

## I. INTRODUCTION

Language learning has never been more important than it is today. Since the onset of globalization, language learning has become essential in facilitating communication across cultures and opening up numerous educational and professional opportunities [6]. To excel in any language, it is crucial to develop proficiency in all four core skills: listening, writing, reading, and speaking. Educational strategies have evolved significantly over the years, transitioning from traditional, lecture-driven methods focused on memorization to modern pedagogical models such as adaptive learning, blended learning, flipped classrooms, problem-solving learning, and game-based learning. These approaches engage learners in active, continuous interaction, improving the overall learning experience and catering to diverse learning styles. In the context of self-directed learning, these methods require the availability of a wide range of creative and diverse pedagogical activities to support personalized learning paths.

The integration of Artificial Intelligence (AI) in education has the potential to reshape the learning landscape, particularly within language acquisition. Both open-source platforms and commercial solutions have embraced AI to enhance the educational experience. For example, Moodle, a widely used Learning Management System (LMS), has incorporated numerous AI-driven features in its latest release, version 4.5 [1]. Similarly, Cypher Learning[2] is leveraging AI technologies to refine the learning process. The advent of AI-based adaptive learning tools has significantly advanced the ability to customize learning journeys, facilitating more effective and personalized educational interventions [?], [?], [13], [14]. Although AI has made notable strides in educational contexts, especially for widely spoken languages, there is still a significant disparity in the development of advanced AI tools for languages with limited resources such Arabic.

Arabic, one of the six official languages of the United Nations, holds immense cultural and historical significance. Spoken by more than 422 million people across 27 countries, it is the fifth most spoken language globally[3]. Despite its widespread use, Arabic remains a low-resourced language, facing significant challenges due to limited access to learning materials and tools. This scarcity complicates the process of acquiring proficiency in Arabic, for both natives and non-natives, when compared to languages like English, Mandarin, Spanish, and Hindi, which benefit from a wealth of educational resources [25]. This highlights the need for the development of

Identify applicable funding agency here. If none, delete this.

---

[1]https://moodle.org/
[2]https://www.cypherlearning.com/
[3]https://www.statista.com/statistics/266808/the-most-spoken-languages-worldwide/

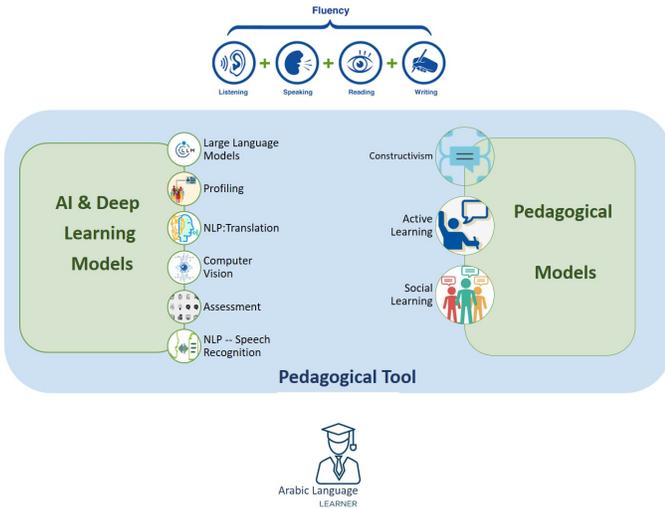

Figure 1: Smart Educational Tool Project

AI-based educational applications tailored to Arabic languages learners.

In response to this need, we launched the Smart-Arabic-Learning project, which focuses on enhancing Arabic language education for non-native speakers. Figure 1 illustrates the salient features of the entire project. This sustainable initiative aims to leverage advancements in Deep Learning models and Active Learning paradigms. The project's core objective is to develop AI-driven educational tools that enrich the Arabic learning experience by offering customized, context-aware activities. Key features include real-world learning scenarios, personalized learning pathways, remediation strategies, and feedback mechanisms, all powered by optimized Deep Learning models to promote diverse pedagogical approaches.

The twofold objectives of this current investigation within this whole project are 1) exploring the potential of Large Language Models (LLMs) and especially Arabic LLM for Visual Question Answering as foundation for educational tools that support the constructivist paradigm, with a focus on proposing visual activities to facilitate Arabic language learning for non-native learners. 2) to bridge the research to market gap in Arabic LLM Deployment. In fact, we aim ease the transition from Arabic LLM research to industrial applications.

This paper is structured as follows. Section II provides the background information, introducing key concepts such as Visual Question Answering (VQA), Visual Question Generation (VQG), and Multimodal Language Models (MLM), which form the foundation of our work. In Section III, we review related work in the field of AI-based educational tools and VQA systems, highlighting their relevance to our research. Section IV outlines our proposed approach, describing the architecture and components of our VQA-based Arabic Educational tool, which integrates VQA and Question Generation (QG) to enhance Arabic language learning. In Section V, we present the experimental setup and evaluation, detailing the benchmark and performance metrics used to assess the tool's effectiveness. Finally, we conclude by summarizing the results and discussing potential future work in this area.

## II. BACKGROUND

Before getting into our work, we start by explaining the fundamental understanding related to Image Captioning and Visual Question Answering (VQA) which are two important tasks in computer vision and natural language processing (NLP) that combine visual information and textual understanding. In addition, a glance at Language Multimodal Models (LMM).

### A. Visual Question Answering

Visual Question Answering (VQA) is a task where a machine answer a natural language question related to an image [4]. VQA combines Computer Vision, Natural Language Processing, and Knowledge Representation and reasoning [28].

Deep Learning have shifted VQA from traditional methods to more sophisticated techniques that improve performance, scalability, and reliability [7], [12], [23]. The key innovations include attention mechanisms, which enhance feature extraction by focusing on specific image regions, and transformer models, which use self-attention to capture long-range dependencies, offering superior modeling of complex relationships. Models like the Vision Transformer (ViT) have extended transformer architecture to visual tasks, improving cross-modal integration.

Additionally, Vision Language Pre-training (VLP) which combines computer vision and natural language processing? involves the development of models and techniques that can process both visual (image or video) and textual (natural language) data simultaneously [7]. The models are trained on large datasets to create robust representations that can be fine-tuned for specific tasks.

Regarding architecture, recent VQA models use advanced CNN-BERT and VLP-based systems. The CNN-BERT architecture leverages BERT for textual encoding and CNN for visual encoding, using transformer-based models like multimodal and crossmodal transformers to better integrate visual and textual data. VLP-based architectures further enhance this by using either single-stream or dual-stream fusion to combine multimodal inputs, allowing for more complex cross-modal interactions. These systems are either encoder-only, generating outputs directly from cross-modal embeddings, or encoder-decoder models, which use a decoder for more refined output generation.

VLP systems are exemplified by pioneer frameworks such as CLIP (Contrastive Language-Image Pretraining) [24], BLIP (Bootstrapping Language-Image Pretraining) [19], and LLaMA (Large Language Model Meta) [31].

Collectively, these approaches enhance the capability of VQA systems to address complex reasoning tasks and process multimodal inputs effectively. Several notable Large Multimodal Models (LMM) have been emerged targeting Arabic, particularly in sentiment analysis, text recognition, and language generation. These include **Qalam** [3], a culturally aware model trained on English-based datasets to represent Arabic dialects, **Dallah** [2], a dialect-aware Arabic LMM, **Fanar** [29] an Arabic-centric platform supporting language generation and speech-to-text recognition, with a focus on cultural preferences in the MENA region, and **AIN** [10], a bilingual English-Arabic model, that covers diverse domains like medical imaging and remote sensing.

*B. Visual Question Generation*

Visual Question Generation (VQG) a compelling issue that has garnered recent attention, involves creating questions based on images, making it more cognitively demanding than question answering. The process requires two key steps [20]: interpreting the image and then generating coherent, valid questions. Image interpretation involves tasks such as object detection, classification, labeling, and understanding relationships between objects and the overall scene. To address this, methods typically use Convolutional Neural Networks (CNNs) for image representation, followed by Recurrent Neural Networks (RNNs) or similar architectures to generate word sequences for the questions.

Most of the studies mentioned for both VQA and VQG use common training or benchmark datasets, such as COCO-QA, and DAQAUR, which contain images representing everyday life. Additionally, there are specialized datasets for VQG, like VQG Flickr-5000 and VQG COCO-5000. However, applying VQA or VQG techniques to specialized domains, such as medical imaging, satellite imagery, and some educational contexts like language learning in under-resourced settings, remains a significant challenge due to the lack of well-curated, domain-specific datasets [1], [15].

## III. Related Work

The integration of Artificial Intelligence (AI) in education is a rapidly growing field, offering a wide range of applications, from foundational learning for young students to professional development and high education [32]. AI systems, particularly those integrating Question-Answering activities, serve as crucial tools for reinforcing and assessing acquired knowledge. Visual Question Answering (VQA) systems, in particular, are valuable educational aids [25].

For instance, He et al. [9] proposed an AI Robot system that aids preschoolers in learning metacognition and geometrical thinking through contextual teaching. The system uses Faster R-CNN to detect objects in a scene and generate questions based on the objects' qualities and colors, successfully engaging children and parents. Sophia and Jacob [27] developed a Student Chatbot during the pandemic, which used RNN for text and CNN for visual tasks, aiding students by addressing their queries. Dipali et al. [17] introduced a VQA system utilizing semantic knowledge from images, combining ResNet for visual features and BERT for textual features. Gyeong-Geon and Xiaoming [18] proposed a VQA system using GPT-4, enhancing the educational value of AI in learning environments.

Concerning Arab world educational Tools some efforts have been made exploring AI usage [**?**], [22], [26].

Sammoudi et al. [26] described the development and evaluation of a textual question-answering model for a personalized learning assistant, using BERT transformers customized for Arabic. The model was fine-tuned on Palestinian 11th and 12th grade biology textbooks, improving its accuracy in answering questions and extracting relevant information in science education.

Mahgoub et al. [22] examine the use of LLMs for textual QA tasks for Arabic languages, specifically for mathematical problems. Despite the models' strong performance in general, they require further fine-tuning to handle mathematical problem-solving effectively. To address the lack of high-quality datasets, the authors propose a Synthetic Data Augmentation framework, which uses LLMs to generate synthetic mathematical problem datasets, enhancing training for QA tasks.

These studies highlight AI's growing role in education, particularly through advanced question-answering systems that foster engagement, curiosity, and deeper learning. While highlighting the fewest investigations that focused on recent VQA usage especially for Arabic language learning [16].

## IV. Approach

Our proposed educational tool, *VQA-ARABIC-EDU*, integrates *Image Captioning* and *Quiz Generation* components to automatically create AI-driven Vision Quizzes. This tool aims to foster active learning by providing learners with various, generous and customized pedagogical activities. The focus is on leveraging the integration and benefits mediation-based learning to enhance learner engagement and effectiveness in the learning process.

Figure 2 depicts the system architecture and its key components. The user-friendly front-end of a web-based application where learners engage in a variety of interactive activities designed to maintain their active participation and targeting a seamless and engaging learning experience [4] . The back-end serves as the core of the system, responsible for generating quizzes built upon image content. During activities such as "VisionQuiz" or "ImageIQ," learners upload images to the platform. These images are processed by our model using deep learning

---
[4]https://github.com/anonymous-for-blind-review-purpose

techniques to extract meaningful features and generate descriptive captions. Subsequently, a language generation model leverages these descriptions to produce diverse and contextually relevant questions, thereby underpinning the visual question answering (VQA) process:

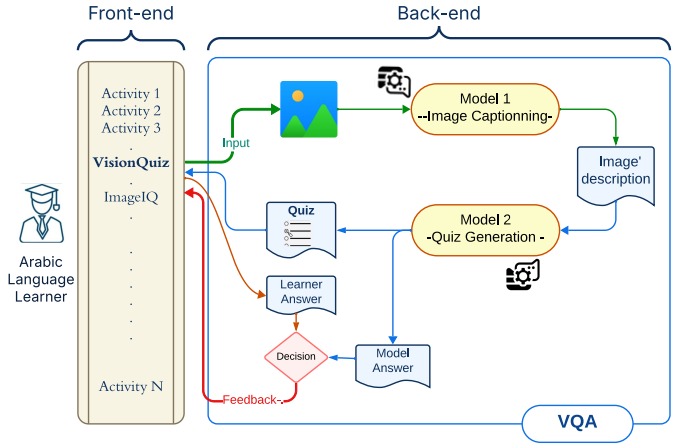

Figure 2: End-to-end Educational Tool Architecture: VisionQuiz Activity

- **Image captionning:** *Model 1* derives textual captions from the provided images. It translates visual data into concise textual summaries using deep learning techniques. A vision-language model is ideal for this purpose, as it supports the pedagogical goal of understanding visual content. In fact, Model 01 identifies key objects, scenes, and spatial relationships within the uploaded image. It converts this information into text, creating clear and simple description that summarize the image content. Any VLP model can be plugged and deployed for this image captioning task. The output description is concealed from the learner.
- **Quizz Generation :** *Model 2* generates diverse and contextually relevant quizzes based on the image description provided by *Model 1*. These questions form the foundation for meaningful interactions within the VQA process.

As the VQA part is responsible for generating quizzes from the image, here is in detail how it works:

1) An image is uploaded into the system. It is saved in a database for easy retrieval, which can then be used for feature extraction, descriptions, or tracking purposes. Learners have the option to select an existing image or randomly upload an image from permitted websites through the VisionQuiz and ImageIQ activity interfaces respectively. The latter option allows learners to freely choose an image, enhancing learner motivation. *Model 1* generates a description of the uploaded image, which is then fed into *Model 2*. This second model generates a set of quizzes based on that description. It is worth mentioning that the question is generated in the user's native language (e.g, English), while the multiple-choice answer' options are provided in Arabic.
2) The learner selects a random question from the generated set of questions delivered by *Model 2*, allowing him to pick the one he prefers to maximise his motivation. In this step, learners respond to questions presented in their native language about images, selecting their answers from multiple-choice options provided in Arabic.
3) **Feedback:** By engaging with these types of questions, learners develop their language skills and progress toward the targeted learning objectives. Upon completion of each task, learners receive feedback based on their responses.

| Task | Prompt Text |
|---|---|
| Image Description | Describe the image accurately and concisely. Focus on objects, actions, people, animals, colors, verbs, adjective. Avoid mentioning brand names. The response must be a single, clear paragraph in english... |
| Question Generation | You are an expert Arabic language tutor helping students learn new vocabulary using visual context. You will be given a description of an image. Your task is to generate exactly two beginner-to-intermediate level multiple-choice questions that help the learner identify Arabic vocabulary related to: - Actions (e.g., walking, eating) - Objects (e.g., table, book) - Colors (e.g., red, green) - Adjectives (e.g., big, small, fast). Each question should:<br>1. Be written in English.<br>2. Ask the learner to choose the correct Arabic word (with full diacritics).<br>3. Offer 4 answer choices in **Arabic** (a, b, c, d).<br>4. Indicate the correct answer clearly. Only return the questions and answers in this format:<br>– Question 1: What is the boy doing?<br>a) يَكْتُبُ b) يَجْلِسُ c) يَأْكُلُ d) يَشْرَبُ  Correct answer: a) يَكْتُبُ<br>– Question 2 : What color is the book?<br>a) أَحْمَرُ b) أَزْرَقُ c) أَخْضَرُ d) أَصْفَرُ  Correct answer: b) أَزْرَقُ<br>Do NOT explain or translate anything. Only output the questions and answers in the format above. |

Table I: Sample of used Prompts.

Our target audience comprised mid-beginner non-native learners. Questions were presented in their native language, while answer choices were given in Arabic. This approach can also serve as a teaching aid to help develop course materials and assignments.

Prompt engineering techniques were employed to generate more precise, language-learning-oriented image descriptions and questions. Table I presents examples of the prompts used for both the image captioning and question generation tasks. These prompts provide detailed contextual information and illustrative examples within the framework of prompt engineering, although the prompting approach is simple and not extensively refined.

## V. EXPERIMENTS

To assess the performance of our VQA-based part within the educational tool, we conducted separate evaluations of its two core modules: image captioning and quiz generation. The evaluation was primarily manual and carried out by four in-lab students following a predefined set of guidelines. All of them are Arabic native speakers. Multiple criteria were considered, including the quality of the generated image descriptions and the relevance of the quizzes in relation to the Arabic language learning objectives.

| Sample (Type, #) | Generated Description (Llama, Gemma) | Generated Quizzes (Llama, Fanar) |
|---|---|---|
| 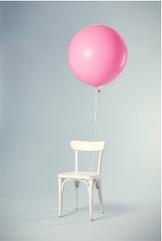 Simple (87) | The image features a white chair with a large pink balloon tied to it, set against a light blue background. The chair is positioned centrally in the frame, facing forward, and has a simple design with a flat backrest and four straight legs. The balloon is inflated to an unusually large size, almost as big as the chair itself, and is secured to the chair by a thin white string that passes through a small hole at the top of the balloon's neck. The background of the image is a plain light blue color, which provides a clean and neutral backdrop for the chair and balloon. Overall, the image presents a whimsical and playful scene, with the oversized balloon adding a touch of humor and surprise to the otherwise mundane setting. | **Q 1**: What color is the balloon? a) أَحْمَرُ (b) أَزْرَقُ (c) وَرَدِيٌّ (d) أَصْفَرُ Correct answer: c) وَرَدِيٌّ **Q 2**: What is attached to the balloon? a) طَاوِلة (b) قَلَم (c) خيط (d) كُرْسِيٌّ Translation ( Table, Pen, String, Chair) Correct answer: c) خيط **Q 3**: What is the chair made of? a) مِعْمَارِي (b) حَدِيد (c) خَشَب (d) بَلَسْتِك Translation ( Architectural, Iron, Wood Plastic) Correct answer: c) خَشَب **Q 4**: In which direction does the string go? a) للأَسْفل (b) للأَعْلى (c) إلى اليَسار (d) إلى اليَمين Translation ( Down, up, left, right) Correct answer: b) للأَعْلى |
| 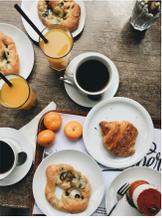 Moderate (56) | A wooden surface displays a breakfast arrangement featuring several white plates holding various pastries: a flaky croissant, a muffin with blueberries, a circular pastry studded with nuts, and a small cake with white filling and reddish-brown coating. Two glasses of orange juice, each with a dark straw, and two cups of dark coffee sit among the food, accompanied by three small, bright orange citrus fruits. A clear glass containing packets rests to the side, and a small silver spoon lies near the cake, suggesting someone has begun eating. | **Q 1**: What action does the person seem to have started? a) القراءة (b) الطبخ (c) الأكل (d) الكِتابة Translation ( Reading, cooking, Eating, Writing ) Correct answer: c) الأكل **Q 2**: Which object do you see that indicates a drink? a) عصير برتقالي في زجاجة (b) كيسات بلاستيكية (c) كوب قهوة داكنة (d) ملاعق فضية صغيرة Translation ( Plastic bags, Orange juice in a bottle, A cup of dark coffee, Small silver spoons Correct answer: c) كوب قهوة داكنة **Q 3**: What is the color of the pastries with nutty toppings? a) أصفر (b) أحمر (c) بني محمر (d) أبيض Correct answer: c) بني محمر |
| 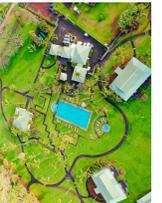 Complex (68) | This aerial image presents a bird's-eye view of a large, sprawling property featuring multiple buildings and a large swimming pool. The property is divided into distinct sections by winding black paths that traverse the lush green grass. The buildings are predominantly white or light-colored, with several sporting green roofs, while the pool area boasts a vibrant blue hue. Scattered throughout the property are numerous palm trees, adding to its tropical ambiance. The overall atmosphere of the image exudes luxury and relaxation, suggesting that this could be a high-end resort or hotel. The presence of multiple buildings and amenities implies that the property is designed to accommodate a significant number of guests. | **Q 1** (Actions): What activity can people do by the large pool? a) القراءة (b) السباحة (c) الجري (d) المشي Correct answer: b) السباحة **Q 2** (Objects): What type of feature surrounds the larger pool? a) طاولات (b) أكواخ (c) مقاعد الاسترخاء (d) مطاعم في الهواء الطلق Correct answer: c) مقاعد الاسترخاء **Q 4**: What is surrounding the pool? a) كُرْسِيٌّ (b) شَجَرَةٌ (c) مَنْزِلُ (d) نَوافِذ Correct answer: b) شَجَرَةٌ |

Table II: Sample of images with their related generated Description and Quizzes

The current evaluation aims mainly assessing both the impact of prompting on the system performances and the quality of the generated quizzes. Let us first describe the test bed.

*A. Deployed LLM*

We have deployed a panoply of LLMs through *Groq*[5] and *kluster.ai*[6] model providers. For the vision, we have used :

- **Llama 3.2-90B Vision**: is a large-scale multimodal language model that integrates visual and textual understanding. It processes images and text jointly, enabling advanced vision-language tasks such as image captioning and visual question answering [8].
- **Gemma 3 27B It**: is a large-scale language model that excels at various NLP tasks such as text completion, summarization, and question answering, leveraging extensive pretraining on diverse multilingual datasets to ensure versatility and accuracy [30].

[5]https://groq.com/
[6]https://www.kluster.ai/

Concerning the quiz generation, we used :

- **Llama 3.3-70B Versatile**: developed to advance natural language understanding and generation [5]. Leveraging extensive pretraining on vast and diverse multilingual corpora, it aims to excel at complex tasks such as contextual reasoning, summarization, and dialogue generation.
- **Fanar** is developed to support natural language understanding and generation tasks with a focus on Arabic language processing [29]. It is optimized for instruction-following and conversational applications, enabling context-aware and accurate responses. FANAR leverages extensive pretraining on diverse Arabic corpora to enhance its linguistic capabilities and provide effective language support in educational and practical domains.

## B. Image Benchmark

We utilize the Unsplash platform [7] to input images into our VisualQuiz system. This website designed for browsing and downloading high-quality, royalty-free images for a wide range of purposes. The platform features an intuitive search function that facilitates the discovery of images based on specific keywords or themes. In this study, we leverage Unsplash to acquire three distinct types of images for our benchmark and to direct the search toward images that hold potential value for language learning.

In fact, to obtain a more reliable evaluation, we collected 211 images categorized into three levels of complexity: Simple, Mid-Complex and Complex. Simple Images have Minimal details, few colors, basic shapes (e.g., icons, logos, line drawings, common objects). Mid-Complex Images – Moderate Complexity – have more details, multiple colors, recognizable objects but not overly intricate (e.g., simple infographics, structured diagrams, minimalistic illustrations). Highly Complex Images have high details, many overlapping elements, intricate textures (e.g., detailed landscapes, medical scans, intricate paintings). These images were randomly uploaded from that dataset.

Table II reports illustration of images by category.

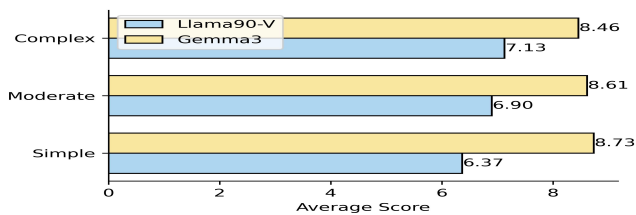

Figure 3: Comparison of image description qualities – Average Scores-: Gemma3 vs Llama90-V by Image Category

[7]https://unsplash.com

## C. Image Description Quality & Effect of Prompting

To assess the performance of the image captioning module, we conducted a manual evaluation. Four native experts evaluated the subset of image descriptions generated by the *Llama 3.2-90B Vision* (Llama90) and *Gemma 3 27B It* (Gemma3) large vision-language models, considering both prompted and non-prompted generation scenarios. A sample of these description are reported in Table II. A total of 844 descriptions were assessed.

The evaluation scale ranged from 0 to 10 and took into account factors such *1) Accuracy and Relevance* : check if the caption correctly describes the main objects, actions, and context present in the image without hallucinations. *2) Completeness* where we check if the caption provide enough detail to convey the essential content of the image- and *3)Alignment with language learning goals* -If the captions are for language learning purposes, and are tailored to the learner's level and objectives.

For this evaluation, a set of guidelines was provided and explained to the annotators. The mean score was used for evaluation, including only the experts' assessments that exhibited a deviation of no more than 2 points.

Figure 3 presents the evaluation scores of image descriptions generated by both Llama90-V and Gemma3. The initial observation indicates that both vision-language models successfully produced acceptable image descriptions, each achieving mean scores above 6.37/10. Notably, Gemma3 outperformed Llama90-V in the quality of generated descriptions, particularly for simple as well as moderately complex images. These results suggest that Gemma3 is well-suited for usage in educational contexts.

Figure4 illustrates more details on the evaluation scores within prompting and no-prompting scenarios while considering the category of images.

The initial insight indicates that prompting does not significantly impact the performance of Llama90-V, whereas it appears to have a more noticeable effect on Gemma3. This difference may be due to variations in the models' architectures, training data, or inherent capabilities. For instance, Llama90-V might be less sensitive to prompt modifications, while Gemma3 may better leverage prompt information to improve output quality.

The low scores (below 4 out of 10), representing 3% for Llama90-V and 0% for Gemma3, typically reflect instances of hallucinations or deficiencies in counting and reasoning. Below are some illustrative examples:

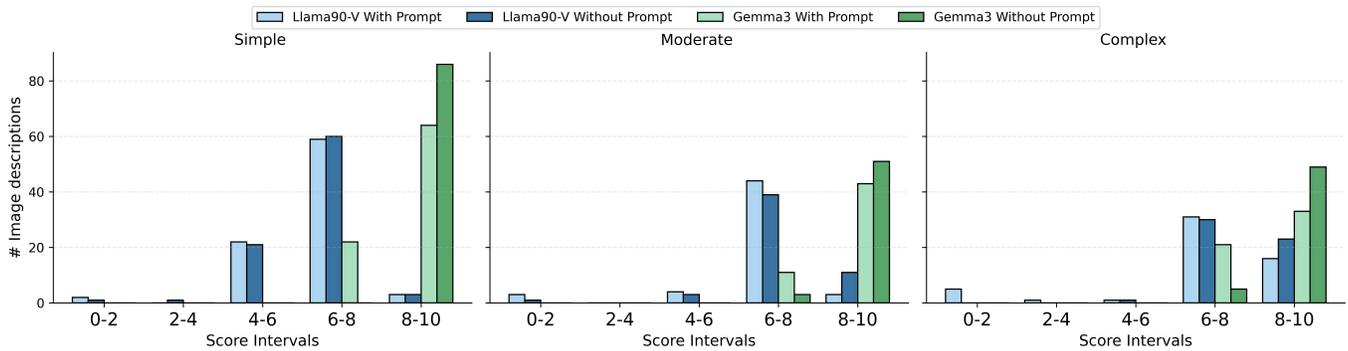

Figure 4: Manual Scoring Distribution of Generated Image Descriptions by Prompt Condition

| Image | Related Description | Issue |
|---|---|---|
| | … **two smooth, white oval objects**. Strong sunlight … …yellow mug filled with milk and **four white eggs** arranged | -Lack of precision -Problem of counting |
| | The image presents a French-language infographic illustrating the distribution of funds for a medical consultation… | Hallucination |

To mitigate such issues, two approaches can be considered. The first involves integrating a human-in-the-loop to manually review and discard problematic outputs—though this solution is often costly and time-consuming. The second approach involves deploying an automated evaluation system designed to assess the accuracy of generated descriptions. In this context, metrics such as CLIPScore [11] and VQAScore [21] can be effectively utilized to quantitatively measure the alignment between the generated textual descriptions and the corresponding visual content.

### D. Quality of the Generated Quizzes

To assess the quality of the generated quizzes, three quizzes per image and per category were produced by each deployed large language model — namely Llama3.3-70B (Llama70) and Fanar. Only those generated by the best-reached image description are assessed totalling 1266 Quizzes. Concerning the question types, we have opted for multi-choices. A sample of these quizzes is presented in Figure II. The human evaluation primarily focused on assessing the alignment of each question with the image description context, as well as the quality of the multiple-choice format, in accordance with Arabic language learning objectives.

The evaluation scale ranged from 0 to 10 and took into account factors such as question complexity, vocabulary richness, and precision. Low scores corresponded to very simple questions using common vocabulary, whereas higher scores reflected more sophisticated questions that required learners to engage in deeper cognitive effort, featuring richer vocabulary, longer expressions, and grammatically well-constructed items.

Furthermore, the assessment incorporated a precision metric based on the clarity and correctness of the multiple-choice answers. Scores were assigned as follows: low scores when the answers caused confusion for learners, moderate scores when answer choices were very similar—considering the learners' mid-beginner proficiency level, and high scores (up to 5 points) when both the question and answer' options were accurately formulated without errors or ambiguity. The provided correct answer is also assessed.

All detailed guidelines were clearly communicated to the annotators to ensure consistent evaluation. Preliminary evaluations were systematically performed to confirm the consistency and precision of the annotation process.

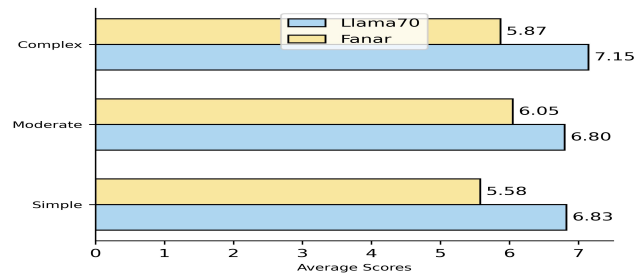

Figure 5: Manual Evaluation of Generated Quizzes: Llama70 vs Fanar

Figure 5 illustrates the manual evaluation scores of the generated quizzes by both Llama70 and Fanar LLMs. // The first observation is that both LLMs achieve above-average scores, surpassing the mean value of 5.58 out of 10. In comparison, Llama70 outperforms Fanar by approximately 22.4% on simple images, 12.4% on moderate

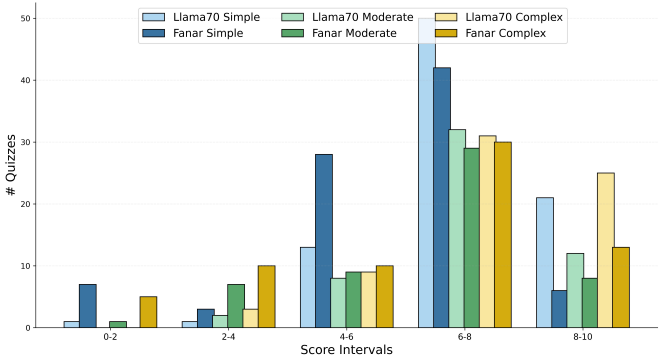

Figure 6: Distribution of manual evaluation scores of Generated Quizzes

images, and 21.8% on complex images.

Figure 6 reports the distribution of manual evaluation through different score intervals by image category.

For high scores (8–10), Llama70 clearly outperforms Fanar in the quality of generated quizzes. However, within the mid-range scores (4–8), both models perform comparably for moderate and complex image categories. Fanar shows particularly strong performance on simple images and dominates the 6–8 score range overall, suggesting better detail retention and more effective mid-level inference.

However, it should be noted that Fanar shows competitive results in mid-range scores and tends to maintain better precision in certain linguistic aspects, such as diacritisation, which is crucial for Arabic language learning especially for non-native learners.

Consequently, for precision-level of language learning such as reasoning and analytic quizzes, Llama70 is the preferred choice due to its superior high-end performance. On the other hand, Fanar may be better suited for balanced where generalization is important, as it tends to exhibit fewer mid-tier failures.

Regarding the low scores (in the 0-4 range), they represent 6% and 8% of the total generated quizzes for Llama70 and Fanar, respectively. We conducted a detailed analysis of the common issues under lying these lower performances. The primary challenges observed include hallucination and ambiguity , particularly arising from multiple choice options that are very closely related in content. Below, some illustrative examples Here some illustration.

| Image | Generated Quiz | Issue |
|---|---|---|
| 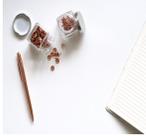 | Q: What are scattered around the jars?<br>a) دَوَاعِع (b) أَقْلَام<br>مُخَطَّطَات (d) مَفَاتِيح (c)<br>Correct answer: b) دَوَاعِع | Failed to give the correct answer دَوَاعِع is not an Arabic word |
| 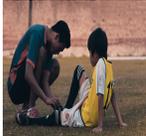 | Q: What is the adult doing to the boy?<br>a) يَصُومُ b) يُشجِعُ c) assin<br>d) يَجري is not correct, instead: y جَري is not a suitable answer, a more suitable answer would be: يَفحَص<br>Correct answer: a) يُشجِعُ is not correct, instead: y فَحَص | - Hallucination when presenting "multi-choice options" and "correct answer"<br>- Diacritisation errors<br>- Code Switching |
| 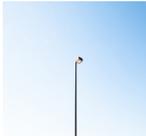 | Q: What does the image show in the middle?<br>a) كَنيوسة طريق (streetlight )<br>b) سماء زرقاء (sky )<br>c) عمود طويل (pole )<br>d) رأس مستطيل (lamp head )<br>Correct answer: a) كَنيوسة طريق | ambiguous options<br><br>Unusual vocabulary |

To check that the system provides Arabic language learners with effective instructional feedback, we manually measured the rate of automatically generated correct answers. The results, presented below, show that out of 1266 generated quizzes, 77.24% included correct answer options.

| Type of Image | Rate (%) |
|---|---|
| Simple | 88.21 |
| Moderate | 76.36 |
| Complex | 67.16 |
| **Global Rate** | **77.24** |

As expected, the simpler the image description, the higher the rate of correct answers provided, suggesting that simple mediation images are more effective in conveying linguistic knowledge.

## VI. Conclusion

We presented an AI-powered educational tool designed to enhance Arabic language learning for non-native speakers through the integration of Visual Question Answering (VQA). By leveraging advanced deep learning models, the tool generates interactive visual quizzes and assignments, providing a contextually rich and engaging learning experience. The use of Vision-Language Pretraining (VLP) models and Large Language Models (LLMs) enables the creation of personalized language learning activities that align with constructivist educational paradigms.

We demonstrated that the tool effectively bridges the gap in Arabic language education, offering learners a promising, AI-driven resource. The results show promising performance in terms of both image captioning and quiz generation, highlighting the potential of this system to support active learning of Arabic language.

Future work will focus on enhancing the tool by incorporating images directly related to the learning context. Instead of relying on external images, we plan to use academic teaching resources, such as illustrations and collections from scholarly books, to further enrich the learning experience and ensure content relevance. This will improve the tool's alignment with curriculum-specific requirements and facilitate more context-driven learning for non-native Arabic speakers. Furthermore, we plan to conduct a real-world evaluation of the tool with non-native Arabic language learners, which is more effective and necessitates a longer period of follow-up.

In our ongoing work, we are exploring some prompt engineering tips to prevent hallucination phenomenon. In fact, we will experiment Chain-of-Thought Prompting and trying to select perfect Quizzes examples, as a basis for fine-tune a lightweight model to replicate that exact extraction behavior.

## References

[1] Said Al Faraby, Adiwijaya Adiwijaya, and Ade Romadhony. Review on neural question generation for education purposes. *International Journal of Artificial Intelligence in Education*, 34(3):1008–1045, 2024.

[2] Fakhraddin Alwajih, Gagan Bhatia, and Muhammad Abdul-Mageed. Dallah: A dialect-aware multimodal large language model for Arabic. In Nizar Habash, Houda Bouamor, Ramy Eskander, Nadi Tomeh, Ibrahim Abu Farha, Ahmed Abdelali, Samia Touileb, Injy Hamed, Yaser Onaizan, Bashar Alhafni, Wissam Antoun, Salam Khalifa, Hatem Haddad, Imed Zitouni, Badr AlKhamissi, Rawan Almatham, and Khalil Mrini, editors, *Proceedings of The Second Arabic Natural Language Processing Conference*, pages 320–336, Bangkok, Thailand, August 2024. Association for Computational Linguistics.

[3] Fakhraddin Alwajih, El Moatez Billah Nagoudi, Gagan Bhatia, Abdelrahman Mohamed, and Muhammad Abdul-Mageed. Peacock: A family of Arabic multimodal large language models and benchmarks. In Lun-Wei Ku, Andre Martins, and Vivek Srikumar, editors, *Proceedings of the 62nd Annual Meeting of the Association for Computational Linguistics (Volume 1: Long Papers)*, pages 12753–12776, Bangkok, Thailand, August 2024. Association for Computational Linguistics.

[4] Stanislaw Antol, Aishwarya Agrawal, Jiasen Lu, Margaret Mitchell, Dhruv Batra, C. Lawrence Zitnick, and Devi Parikh. Vqa: Visual question answering. *arXiv preprint arXiv:1612.00837*, 2016.

[5] M Saiful Bari, Yazeed Alnumay, Norah A. Alzahrani, Nouf M. Alotaibi, Hisham A. Alyahya, Sultan AlRashed, Faisal A. Mirza, Shaykhah Z. Alsubaie, Hassan A. Alahmed, Ghadah Alabduljabbar, Raghad Alkhathran, Yousef Almushayqih, Raneem Alnajim, Salman Alsubaihi, Maryam Al Mansour, Majed Alrubaian, Ali Alammari, Zaki Alawami, Abdulmohsen Al-Thubaity, Ahmed Abdelali, Jeril Kuriakose, Abdalghani Abujabal, Nora Al-Twairesh, Areeb Alowisheq, and Haidar Khan. Allam: Large language models for arabic and english, 2024.

[6] David Block. *Language Education and Globalization*, pages 31–43. Springer US, Boston, MA, 2008.

[7] Fei-Long Chen, Du-Zhen Zhang, Ming-Lun Han, Xiu-Yi Chen, Jing Shi, Shuang Xu, and Bo Xu. Vlp: A survey on vision-language pre-training. *Machine Intelligence Research*, 20(1):38–56, 2023.

[8] Aaron Grattafiori, Abhimanyu Dubey, Abhinav Jauhri, Abhinav Pandey, Abhishek Kadian, Ahmad Al-Dahle, Aiesha Letman, Akhil Mathur, Alan Schelten, Alex Vaughan, et al. The llama 3 herd of models. *arXiv preprint arXiv:2407.21783*, 2024.

[9] Bin He, Meng Xia, Xinguo Yu, Pengpeng Jian, Hao Meng, and Zhanwen Chen. An educational robot system of visual question answering for preschoolers. In *2017 2nd international conference on robotics and automation engineering (ICRAE)*, pages 441–445. IEEE, 2017.

[10] Ahmed Heakl, Sara Ghaboura, Omkar Thawkar, Fahad Shahbaz Khan, Hisham Cholakkal, Rao Muhammad Anwer, and Salman Khan. Ain: The arabic inclusive large multimodal model. *arXiv preprint arXiv:2502.00094*, 2025.

[11] Jack Hessel, Ari Holtzman, Maxwell Forbes, Ronan Le Bras, and Yejin Choi. CLIPScore: A reference-free evaluation metric for image captioning. In *Proceedings of the 2021 Conference on Empirical Methods in Natural Language Processing*, pages 7514–7528. Association for Computational Linguistics, 2021.

[12] D. A. Hudson and C. D. Manning. Gqa: A new dataset for real-world visual reasoning and compositional question answering. In *Proceedings of the IEEE/CVF Conference on Computer Vision and Pattern Recognition (CVPR)*, pages 6700–6709, 2019.

[13] Fenglin Jia, Daner Sun, Qing Ma, and Chee-Kit Looi. Developing an ai-based learning system for l2 learners' authentic and ubiquitous learning in english language. *Sustainability*, 14(23):15527, 2022.

[14] Tumaini Kabudi, Ilias Pappas, and Dag Håkon Olsen. Ai-enabled adaptive learning systems: A systematic mapping of the literature. *Computers and education: Artificial intelligence*, 2:100017, 2021.

[15] Sarah kamel, Shimaa Hassan, and Lamiaa Elrefaei. Vaqa: Visual arabic question answering. *Arabian Journal for Science and Engineering*, 48, 03 2023.

[16] Soumia Khedimi, Abdelghani Bouziane, and Djelloul Bouchiha. Advancements and challenges in arabic question answering systems: a comprehensive survey. *Brazilian Journal of Technology*, 7(4):e75604–e75604, 2024.

[17] Dipali Koshti, Ashutosh Gupta, Mukesh Kalla, Pramit Kanjilal, Sushant Shanbhag, and Nirmit Karkera. Eduvqa–visual question answering: An educational perspective. *Journal of Advanced Research in Applied Sciences and Engineering Technology*, 42(1):144–157, 2024.

[18] Gyeong-Geon Lee and Xiaoming Zhai. Realizing visual question answering for education: Gpt-4v as a multimodal ai. *arXiv preprint arXiv:2405.07163*, 2024.

[19] Shuang Li, Junnan Li, Yingwei Pan, Weizhe Liu, Bo Li, and Yu Qiao. Blip: Bootstrapping language-image pretraining. *arXiv preprint arXiv:2201.12086*, 2022.

[20] Yikang Li, Nan Duan, Bolei Zhou, Xiao Chu, Wanli Ouyang, Xiaogang Wang, and Ming Zhou. Visual question generation as dual task of visual question answering. In *Proceedings of the IEEE conference on computer vision and pattern recognition*, pages 6116–6124, 2018.

[21] Zhiqiu Lin, Deepak Pathak, Baiqi Li, Jiayao Li, Xide Xia, Graham Neubig, Pengchuan Zhang, and Deva Ramanan. Evaluating text-to-visual generation with image-to-text generation. In Aleš Leonardis, Elisa Ricci, Stefan Roth, Olga Russakovsky, Torsten Sattler, and Gül Varol, editors, *Computer Vision – ECCV 2024*, pages 366–384, Cham, 2025. Springer Nature Switzerland.

[22] Abeer Mahgoub, Ghada Khoriba, and ElHassan Anas ElSabry. Mathematical problem solving in arabic: Assessing large language models. *Procedia Computer Science*, 244:86–95, 2024.

[23] M. M. Naseer, K. Ranasinghe, S. H. Khan, M. Hayat, F. Shahbaz Khan, and M. H. Yang. Intriguing properties of vision transformers. In *Advances in Neural Information Processing Systems (NeurIPS)*, volume 34, pages 23296–23308, 2021.

[24] Alec Radford, Jong Wook Kim, Chris Hallacy, Aditya Ramesh, Gabriel Goh, Sandhini Agarwal, Alethea Power, H. Lee, David Luan, Daniel Y. Fu, and et al. Learning transferable visual models from natural language supervision. *arXiv preprint arXiv:2103.00020*, 2021.

[25] Muhammad Yusuf Salam, Fitri Silva, and Hafifa Zulan. Educational technology based on artificial intelligence in supporting distance learning in arabic language learning in higher education. *JILTECH: Journal International of Lingua & Technology*, 3(1), 2024.

[26] Mohammad Sammoudi, Ahmad Habaybeh, Huthaifa I Ashqar, and Mohammed Elhenawy. Question-answering (qa) model for

a personalized learning assistant for arabic language. In *International Conference on Intelligent Systems, Blockchain, and Communication Technologies*, pages 356–367. Springer, 2024.

[27] J Jinu Sophia and T Prem Jacob. Edubot-a chatbot for education in covid-19 pandemic and vqabot comparison. In *2021 Second International Conference on Electronics and Sustainable Communication Systems (ICESC)*, pages 1707–1714. IEEE, 2021.

[28] Hao Tan and Mohit Bansal. Lxmert: Learning cross-modality encoder representations from transformers. In *Proceedings of the 2019 Conference on Empirical Methods in Natural Language Processing (EMNLP)*, pages 5099–5110. Association for Computational Linguistics, 2019.

[29] Fanar Team, Ummar Abbas, Mohammad Shahmeer Ahmad, Firoj Alam, Enes Altinisik, Ehsannedin Asgari, Yazan Boshmaf, Sabri Boughorbel, Sanjay Chawla, Shammur Chowdhury, Fahim Dalvi, Kareem Darwish, Nadir Durrani, Mohamed Elfeky, Ahmed Elmagarmid, Mohamed Eltabakh, Masoomali Fatehkia, Anastasios Fragkopoulos, Maram Hasanain, Majd Hawasly, Mus'ab Husaini, Soon-Gyo Jung, Ji Kim Lucas, Walid Magdy, Safa Messaoud, Abubakr Mohamed, Tasnim Mohiuddin, Basel Mousi, Hamdy Mubarak, Ahmad Musleh, Zan Naeem, Mourad Ouzzani, Dorde Popovic, Amin Sadeghi, Husrev Taha Sencar, Mohammed Shinoy, Omar Sinan, Yifan Zhang, Ahmed Ali, Yassine El Kheir, Xiaosong Ma, and Chaoyi Ruan. Fanar: An arabic-centric multimodal generative ai platform, 2025.

[30] Gemma Team, Aishwarya Kamath, Johan Ferret, Shreya Pathak, Nino Vieillard, Ramona Merhej, Sarah Perrin, Tatiana Matejovicova, Alexandre Ramé, Morgane Rivière, et al. Gemma 3 technical report. *arXiv preprint arXiv:2503.19786*, 2025.

[31] Hugo Touvron, Louis Martin, Benjamin Lespiau, Thomas Daudé, Gauthier Le Mével, Damien Scialom, Rémi Gabral, Pierre G. J., and et al. Llama: Open and efficient foundation language models. *arXiv preprint arXiv:2302.13971*, 2023.

[32] PS Venkateswaran, Firas Tayseer Mohammad Ayasrah, Varun Kumar Nomula, P Paramasivan, P Anand, and K Bogeshwaran. Applications of artificial intelligence tools in higher education. In *Data-driven decision making for long-term business success*, pages 124–136. Igi Global, 2024.